\documentclass[journal]{IEEEtran}

\ifCLASSINFOpdf
\else
   \usepackage[dvips]{graphicx}
\fi
\usepackage{url}

\usepackage{graphicx}
\usepackage{amsmath}
\usepackage{amssymb}
\usepackage{algorithmic}
\usepackage{algorithm}

\begin{document}

\title{Learning data association without data association: An EM approach to neural assignment prediction}

\author{Michael Burke, and Subramanian Ramamoorthy
\thanks{This research was supported by the Alan Turing Institute, as part of the Safe AI for surgical assistance project. This work has been submitted to the IEEE for possible publication. Copyright may be transferred without notice, after which this version may no longer be accessible.}
\thanks{M. Burke is with Electrical and Computer Systems Engineering, Monash University, Australia, and S. Ramamoorthy is with the Institute of Perception, Action and Behaviour, University of Edinburgh, Edinburgh, UK (e-mail: michael.burke1@monash.edu,s.ramamoorthy@ed.ac.uk).}}

\maketitle

\begin{abstract}
Data association is a fundamental component of effective multi-object tracking. Current approaches to data-association tend to frame this as an assignment problem relying on gating and distance-based cost matrices, or offset the challenge of data association to a problem of tracking by detection. The latter is typically formulated as a supervised learning problem, and requires labelling information about tracked object identities to train a model for object recognition. This paper introduces an expectation maximisation approach to train neural models for data association, which does not require labelling information. Here, a Sinkhorn network is trained to predict assignment matrices that maximise the marginal likelihood of trajectory observations. Importantly, networks trained using the proposed approach can be re-used in downstream tracking applications.
\end{abstract}

\begin{IEEEkeywords}
Data association, multi-object tracking, Kalman filtering, smoothing
\end{IEEEkeywords}

\IEEEpeerreviewmaketitle

\section{Introduction and related work}

Data association aims to match measurements with state estimates as part of a tracking algorithm. This is typically accomplished using gating of some form, which restricts the possible set of measurement assignments to track associations using a suitable distance metric or association probability. Given these assignment probabilities, gating approaches to data association \cite{blackman1986multiple} attempt to associate measurements with tracks using an assignment algorithm within a tracker. 

Assignment can be hard, using the Hungarian algorithm \cite{kuhn1955hungarian} or a global nearest neighbour (GNN) approach \cite{bar1995multitarget}, or softer, for example via the probabilistic data association filter (P-DAF). The P-DAF \cite{bar1975tracking} applies Bayesian filtering to an average measurement formed by weighting by association probabilities \cite{Vo15} for single target tracking. A combination of soft assignment and filtering gives rise to the joint-probabilistic data-association filter (JP-DAF) \cite{bar1990tracking,barshalom09}, with a computational burden that grows exponentially with the number of targets. Multiple hypothesis tracking (MHT) \cite{Reid79, streit1995probabilistic} takes a deferred decision approach \cite{Vo15}, maintaining a tree of potential data associations as part of a tracking filter, and relying on a range of strategies proposed to occasionally prune the tree to avoid combinatorial explosion (eg. clustering as in \cite{Reid79}). 

As an alternative, some random-finite-set-based tracking (eg. the PhD filter \cite{Vo06}) approaches ignore the data-association problem altogether, and simply seek to keep track of collections of objects over shorter time periods, as this is often sufficient for a wide range of applications. In direct contrast, tracking-by-detection \cite{Bewley16,Bergmann2019TrackingWB} approaches seek to offset the problem of data association to a detector, assuming that contextual information is available to help disambiguate targets. 

This letter follows a tracking-by-detection paradigm, but introduces an alternative approach, that trains neural networks to perform data association conditioned on contextual information, by combining filtering and a diferentiable assignment operation that allows backprogation. This self-supervised training mechanism requires no labelling and produces data-association models (or detector networks for tracking-by-detection) that can be re-used in downstream applications. 

\section{Problem formulation}

Without loss of generality, consider a linear Gaussian multi-object tracking problem. Let $\mathbf{x}_k \in \mathbb{R}^{N\times d}$ denote the $d$-dimensional states of $N$ objects at the $k$-th time step. Further, let $\mathbf{z}_k$ denote observations of a permuted $(\mathbf{P}_k)$ subset $(\mathbf{H}_k)$ of these object states, subject to some Gaussian measurement noise, $\mathbf{R}$. Assume that object states evolve according to a linear Gaussian dynamics model $\mathbf{x}_k = \mathbf{F} \mathbf{x}_{k-1}$ subject to process noise, $\mathbf{Q}$,
\begin{IEEEeqnarray}{lCr}
p(\mathbf{x}_k|\mathbf{x}_{k-1}) &=& \mathcal{N}(\mathbf{x}_k|\mathbf{F} \mathbf{x}_{k-1},\mathbf{Q})\\
p(\mathbf{z}_k|\mathbf{x}_{k},\mathbf{P}_k) &=& \mathcal{N}(\mathbf{z}_k|\mathbf{H}_k \mathbf{P}_k \mathbf{x}_{k},\mathbf{R})
\end{IEEEeqnarray}

Given a trajectory $\mathbf{z}_{1:K}$ of $K$ observations, our goal is to identify the underlying states of the $N$ objects, and train a model to predict the permutation $\mathbf{P}_k$ that associates states $\mathbf{x}_k$ with observations $\mathbf{z}_k$.

\section{Maximum likelihood data association}

\subsection{Kalman Filtering}

Given the linear Gaussian dynamics and measurement models above, we can compute a predictive posterior in closed form using a Kalman filter prediction step, conditioned on a series of data associations $\mathbf{P}_{1:k-1}$,
\begin{IEEEeqnarray}{lCr}
p(\mathbf{x}_k|z_{1:k-1},\mathbf{P}_{1:k-1}) = \nonumber\\\int p(\mathbf{x}_k|\mathbf{x}_{k-1})p(\mathbf{x}_{k-1}|\mathbf{z}_{1:{k-1}},\mathbf{P}_{1:k-1}) d \mathbf{x}_{k-1} \\
= \mathcal{N}\left(\mathbf{x}_{k}|\mathbf{F}\mathbf{\mu}_k,\mathbf{F}\mathbf{\Sigma}_k\mathbf{F}^\text{T} + \mathbf{Q}\right) \\
= \mathcal{N}\left(\mathbf{x}_{k}|\hat{\mathbf{\mu}}_k,\hat{\mathbf{\Sigma}_k}\right). \label{eq:pred}
\end{IEEEeqnarray}
Here, $p(\mathbf{x}_{k-1}|\mathbf{z}_{1:{k-1}},\mathbf{P}_{1:k-1})$ is the Gaussian posterior from a Kalman filter update step,
\begin{IEEEeqnarray}{lCl}
p(\mathbf{x}_{k}|\mathbf{z}_{1:{k}},\mathbf{P}_{1:k}) &=& 
\frac{p(\mathbf{z}_k|x_{k},\mathbf{P}_{k})p(\mathbf{x}_k|\mathbf{z}_{1:k-1},\mathbf{P}_{1:k-1})}{p(\mathbf{z}_k|\mathbf{P}_{1:k},\mathbf{z}_{1:k-1})} \\  &=& \mathcal{N}\left(\mathbf{x}_{k}|\mathbf{\mu}_k,\mathbf{\Sigma}_k\right).\label{eq:post}
\end{IEEEeqnarray}

The marginal likelihood is given by
\begin{IEEEeqnarray}{lCr}
p(\mathbf{z}_k|\mathbf{P}_{1:k},\mathbf{z}_{1:k-1}) = \nonumber \\ \int p(\mathbf{z}_k|\mathbf{x}_{k},\mathbf{P}_k) p(\mathbf{x}_k|\mathbf{z}_{1:k-1},\mathbf{P}_{1:k-1}) d \mathbf{x}_{k} \\ = \mathcal{N}\left(\mathbf{z}_{k}|\mathbf{H}_k\mathbf{P}_k\hat{\mathbf{\mu}_k},(\mathbf{H}_k\mathbf{P}_k)\hat{\mathbf{\Sigma}_k}(\mathbf{H}_k\mathbf{P}_k)^\text{T} + \mathbf{R}_k\right),
\label{eq:marg}
\end{IEEEeqnarray}

and a maximum likelihood estimate for the series of data associations can be obtained as 
\begin{IEEEeqnarray}{lCr}
\bar{\mathbf{P}}_{1:K} &=& \text{argmax}_{\mathbf{P}_{1:K}} p(\mathbf{z}_{1:K}|{\mathbf{P}_{1:K}}) \nonumber \\&=& \prod_{k=1:K} p(\mathbf{z}_{k}|\mathbf{P}_{1:k},\mathbf{z}_{1:k-1})\label{eq:MLest}.
\end{IEEEeqnarray}





\subsection{Data association by permutation prediction}

Solving for the set of permutations $\mathbf{P}_{1:K}$ that maximise the marginal likelihood in (\ref{eq:MLest}) rapidly becomes computationally intractable in most settings, given the large number of potential associations. Instead, we propose to predict permutations  $\mathbf{P}_k$ using a suitable neural network $g_\theta(\cdot)$ with parameters $\theta$, which thus allows the use of gradient descent for inference,
\begin{IEEEeqnarray}{lCl}
\mathbf{P}_{k} &\approx& g_\theta(\mathbf{z}_{k}) \\
\theta^* &=& \text{argmax}_\theta \,\,\, \prod_{k=1:K} p(\mathbf{z}_{k}|g_\theta(\mathbf{z}_{1:k}),\mathbf{z}_{1:k-1})\\
\theta^* &=& \text{argmin}_\theta \,\,\, = -\sum_{k=1:K} \log p(\mathbf{z}_{k}|g_\theta(\mathbf{z}_{1:k}),\mathbf{z}_{1:k-1}) \label{eq:ll}
\end{IEEEeqnarray}
where the abuse of notation $g_\theta(\mathbf{z}_{1:k})$ denotes a set of $\mathbf{P}_{1} \dots \mathbf{P}_{k}$ permutations independently predicted using measurements $\mathbf{z}_{1} \dots \mathbf{z}_{k}$ respectively. It should be noted that for the purposes of this derivation, the network is trained to predict a permutation of observations conditioned on the observations themselves, $g_\theta(\cdot)$ = $g_\theta(\mathbf{z}_{k})$, but the proposed approach is not limited to this. Indeed, in most practical applications this network would also be conditioned on additional contextual information relevant to the multi-object tracking task, such as an image or other relevant features.

\subsection{Sinkhorn networks}
Since permutations are typically hard assignment operations, training a network to predict these is non-trivial as the flow of gradient information is inhibited. We address this using Sinkhorn networks \cite{mena2018learning}, which rely on the Sinkhorn operator $S(\mathbf{X})$ \cite{sinkhorn1964}, applied to a square matrix $\mathbf{X}$,  \begin{eqnarray}
S^0(\mathbf{X}) &=& \exp(\mathbf{X})\\
S^l(\mathbf{X}) &=& \mathcal{T}_{col}(\mathcal{T}_{row}(S^{l-1}(\mathbf{X})))\\
S(\mathbf{X}) &=& \lim_{l\to\infty}S^{l}(\mathbf{X})).
\end{eqnarray}
$\mathcal{T}_{col}(\cdot)$ and $\mathcal{T}_{row}(\cdot)$ denote column and row normalisation operations respectively. The Sinkhorn operation repeatedly iterates between column and row-wise normalisation operators, so that the matrix resulting from this operation is a doubly stochastic (rows and columns sum to one) assignment matrix from the Birkhoff polytope \cite{sinkhorn1964}. The extreme points of this polytope are permutation matrices. Mena et al. \cite{mena2018learning} introduce a differentiable approximation to the permutation $\mathbf{P}$ by applying the Sinkhorn operator, \begin{equation}
\mathbf{P} = \lim_{\tau\to0^+} S(\mathbf{X}/\tau),
\end{equation}
to $\mathbf{X} = g_\theta(\cdot)$, a square matrix predicted using a suitable feed-forward neural network. $\tau$ is a temperature parameter that controls how soft the assignment is.

\subsection{An EM approach to joint filtering and data association}

Algorithm \ref{alg} summarises the proposed data association model training process. At each expectation maximisation iteration, state-observation associations are predicted using a Sinkhorn network. Kalman filtering is then applied to compute the posterior over state estimates, and the marginal likelihood. Since closed form solutions exist for the marginal likelihood, this process is fully differentiable, and neural network parameters are then updated by taking a gradient step to increase the marginal likelihood. After $N_i$ iterations the trained data association model $g_\theta(\cdot)$ and state estimates $\bar{\mathbf{x}_{1:k}}$ are returned.

\begin{algorithm}[H]
\caption{Joint data association and state estimation\label{alg}}
\begin{algorithmic}
\renewcommand{\algorithmicrequire}{\textbf{Input:}}
 \renewcommand{\algorithmicensure}{\textbf{Output:}}
 \REQUIRE Observations $\mathbf{z}_{1:k}$, Learning rate $\lambda$
 \ENSURE  $\mathbf{x}_{1:k}$, $g_\theta(\cdot)$
  \STATE $\theta \sim \mathcal{N}(0,1)$
  \FOR {$i = 0$ to $N_i$}
  \FOR {$k = 0$ to $K$}
  \STATE Predict $\mathbf{P}_{k} = g_\theta(\mathbf{z}_{k})$ using a Sinkhorn network
  \STATE Compute $p(\mathbf{x}_k|z_{1:k-1},\mathbf{P}_{1:k-1})$ using (\ref{eq:pred})
  \STATE Compute $p(\mathbf{z}_k|\mathbf{P}_{1:k},\mathbf{z}_{1:k-1})$ using (\ref{eq:marg})
   \STATE Compute $p(\mathbf{x}_k|\mathbf{P}_{1:k},\mathbf{z}_{1:k})$ using (\ref{eq:post})
  \ENDFOR
  \STATE Compute $\mathcal{LL}$ = $-\sum_{k=1:K} \log p(\mathbf{z}_{k}|\mathbf{P}_{1:k-1},\mathbf{z}_{1:k-1})$
  \STATE $\theta = \theta + \lambda\nabla_{\theta} \mathcal{LL}$
  \ENDFOR
 \RETURN State estimates $\bar{\mathbf{x}_{1:k}}$, Trained model $g_\theta(\cdot)$
\end{algorithmic}
\end{algorithm}

Importantly, the trained data association model $g_\theta(\cdot)$ can then be re-used in downstream tracking applications. Moreover, since the permutation prediction process using the Sinkhorn network produces a soft assignment matrix, downstream applications could also make use of more advanced multiple hypothesis tracking approaches (eg. JP-DAF \cite{bar1990tracking}).

Finally, it should be noted that although the derivation above was presented using the Kalman filter, Kalman smoothing \cite{rauch1965maximum} using a forward and backward pass can also be applied to compute the marginal likelihood.

\subsection{Graduated optimisation}

Jointly smoothing and training data association models using the EM approach is something of a chicken and egg problem, and a difficult optimisation objective. However, the proposed approach naturally lends itself to graduated optimisation \cite{thacker1996vision}. 

Here we start the EM algorithm using a substantially reduced process noise covariance $\mathbf{Q}_0 = \mathbf{Q}^{min}$, and gradually increase this with each iteration,
\begin{equation}
    \mathbf{Q}_{i+1} = \gamma \mathbf{Q}_{i},
\end{equation}
using a suitable graduation rate $\gamma$, until it reaches the true dynamics model covariance, $\mathbf{Q}$. This allows the assignment prediction problem to be improved on a heavily smoothed, simpler tracking task, and then refined as additional complexity in the filtered dynamics is introduced.

\section{Experimental results}

The proposed approach is investigated using a synthetic random walk setting and a vision-based point tracking problem.

\subsection{2D Random Walk}

Here, synthetic observations and ground truth trajectories are generated for 4 points in 2D space using the random walk motion model and a Gaussian measurement model,
\begin{IEEEeqnarray}{lCl}
\mathbf{x}_0 &\sim& \mathcal{N}\left(\mathbf{0}, \mathbf{I}\right)\\
\mathbf{x}_k &\sim& \mathcal{N}\left(\mathbf{x}_{k-1}, \sigma_q^2\mathbf{I}\right)\\
\mathbf{z}_k &\sim& \mathbf{P}_k\,\mathcal{N}\left(\mathbf{x}_{k}, \sigma_r^2\mathbf{I}\right).
\end{IEEEeqnarray}
$\mathbf{P}_k$ denotes a permutation corresponding to a random reordering of points. For this synthetic problem, we are able to baseline against a lower bound assignment approach using the Hungarian algorithm.

\begin{figure}
    \centering
    \includegraphics[width=0.4\textwidth]{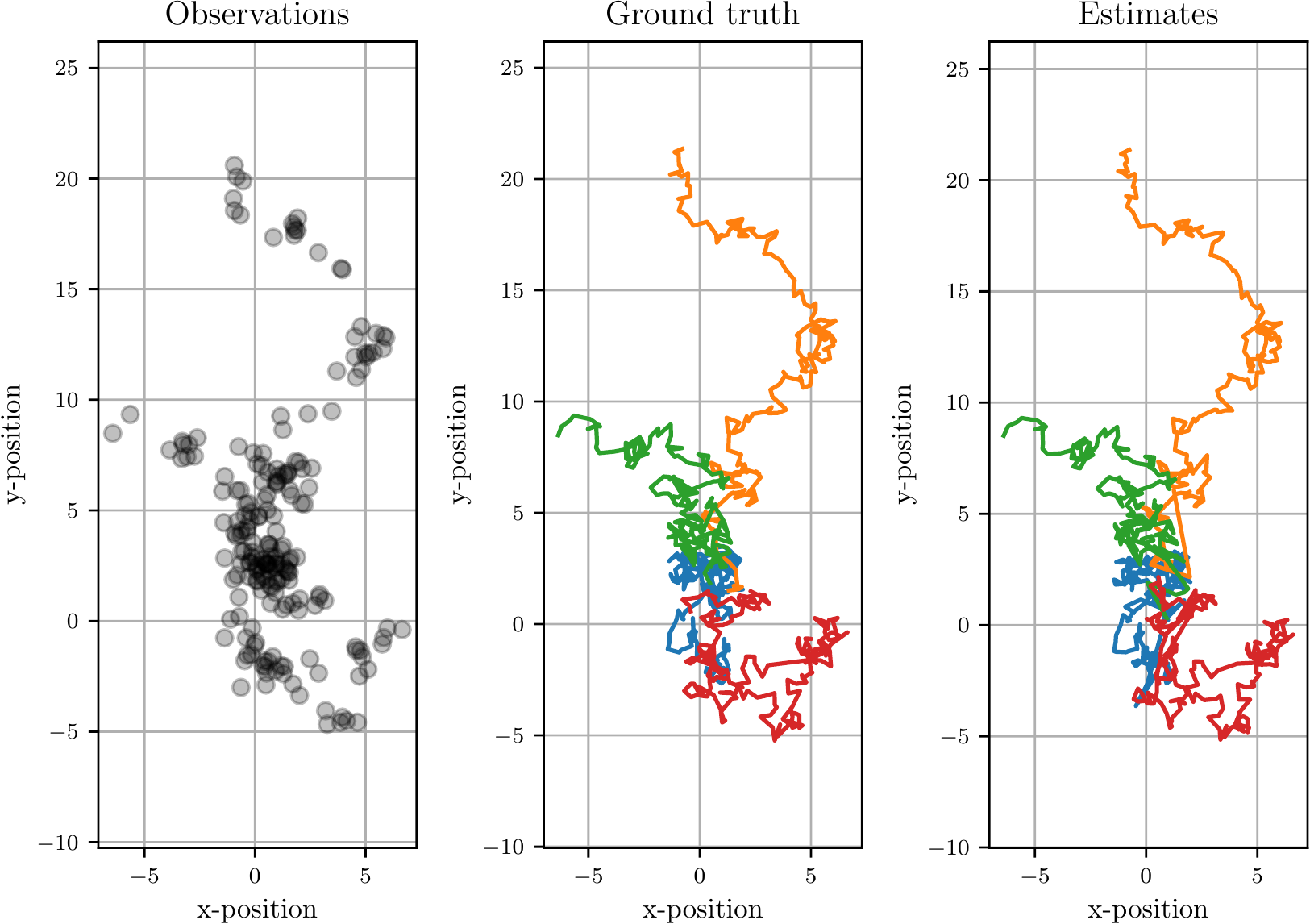}
    \caption{Sample trajectories identified using the joint data association network training and state estimation process. Some association errors occur in particularly challenging regions where observation association is difficult, but, on the whole, the trained neural network is effective at data association.}
    \label{fig:tracks}
\end{figure}

We explore the performance of the proposed approach as a function of measurement uncertainty, learning rates and with or without graduated optimisation. Fig. \ref{fig:tracks} shows smoothed position estimates for a single trial. It is clear that the neural network has learned to correctly associate observations with states, although some incorrect assignments are made in particularly challenging situations.

\subsubsection{Effect of graduated optimisation}

\begin{figure}
    \centering
    \includegraphics[width=0.4\textwidth]{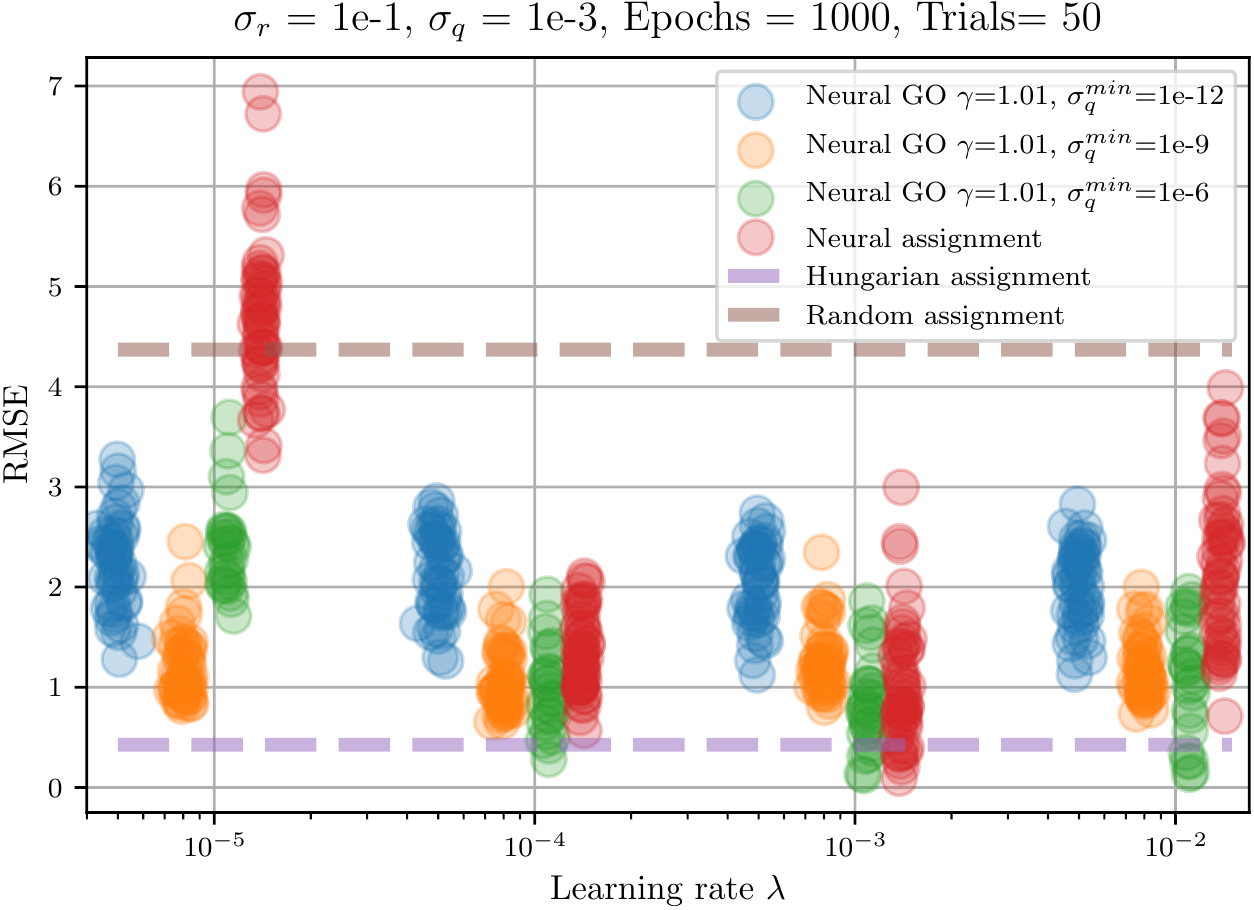}
    \caption{RMSE curves as a function of learning rates and with varying levels of graduated optimisation show that, with appropriate hyper-parameter settings, the neural network can learn to associate data as well as the Hungarian algorithm lower bound.}
    \label{fig:learning}
\end{figure}

Figure \ref{fig:learning} shows the root mean square error (RMSE) in filtered state estimates as a function of the neural network learning rate $\lambda$, with and without varying rates of graduated optimisation (go). Experiments were repeated 50 times with different random seeds. As a lower bound, the Hungarian algorithm is applied to a Euclidean distance cost matrix. In this random walk motion setting, with no clutter, this represents the best case data association.

Results show that, in general, graduated optimisation reduces errors obtained from improperly selected learning rates. With appropriate hyper-parameter settings, the neural network can learn to associate data as well as the Hungarian algorithm. It should be noted that, although the Hungarian algorithm is used as a lower bound for these experiments, it cannot be directly applied to high dimensional, non-positional data, for example images, which is where the primary benefits of training neural networks for data association arise.  

\subsubsection{Robustness to noise}

\begin{figure}
    \centering
    \includegraphics[width=0.4\textwidth]{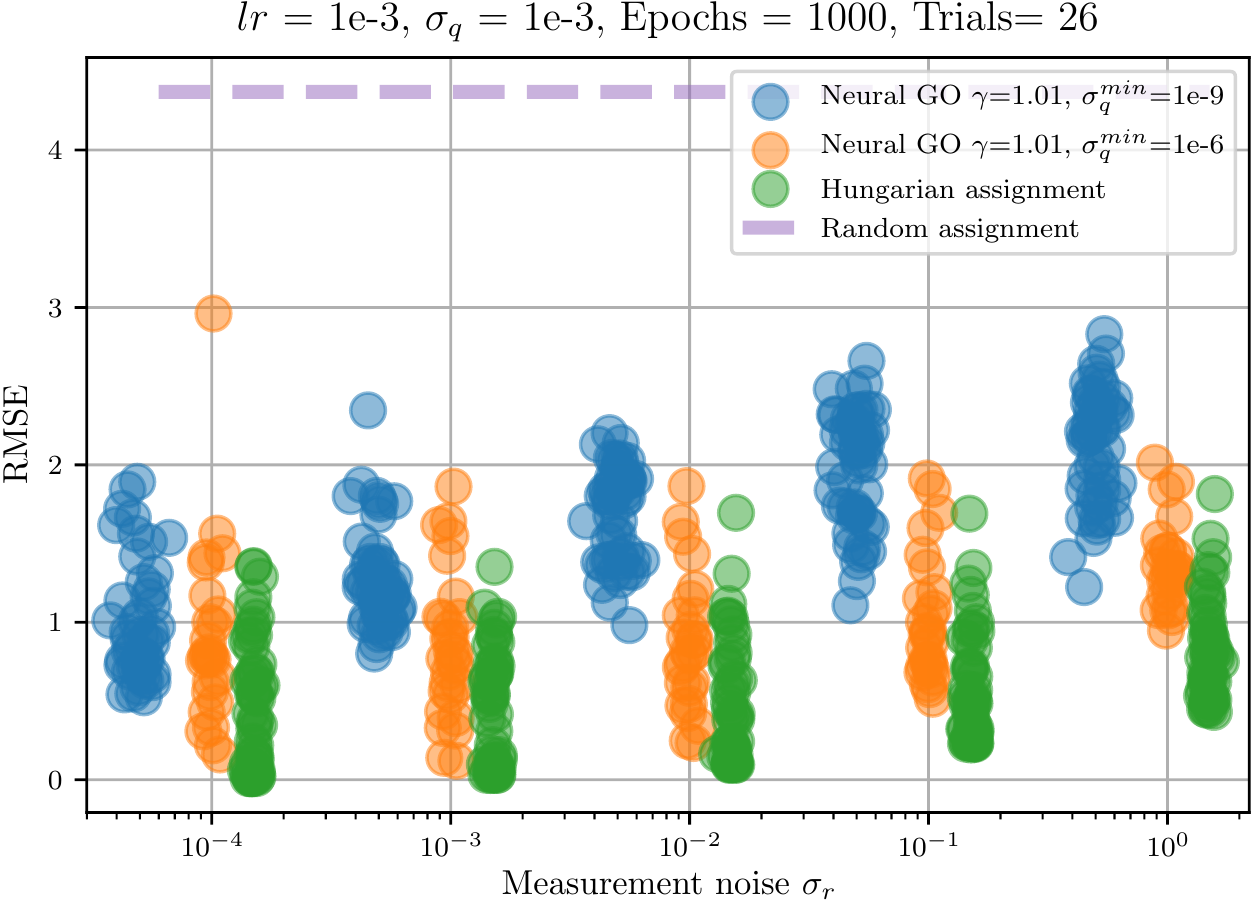}
    \caption{As the SNR is decreased, the neural data association model performance reduces on par with the Hungarian algorithm lower bound.}
    \label{fig:noise}
\end{figure}

Figure \ref{fig:noise} shows the root mean square error as the measurement noise is increased. Smoothing using the Hungarian algorithm on a distance-based cost matrix is included as a baseline. As before, experiments were repeated 50 times with different random seeds. As the SNR is decreased, the neural data association model performance reduces on par with the Hungarian algorithm lower bound.


\subsection{Vision-based object recognition}

\begin{figure}
    \centering
    \includegraphics[width=0.45\textwidth]{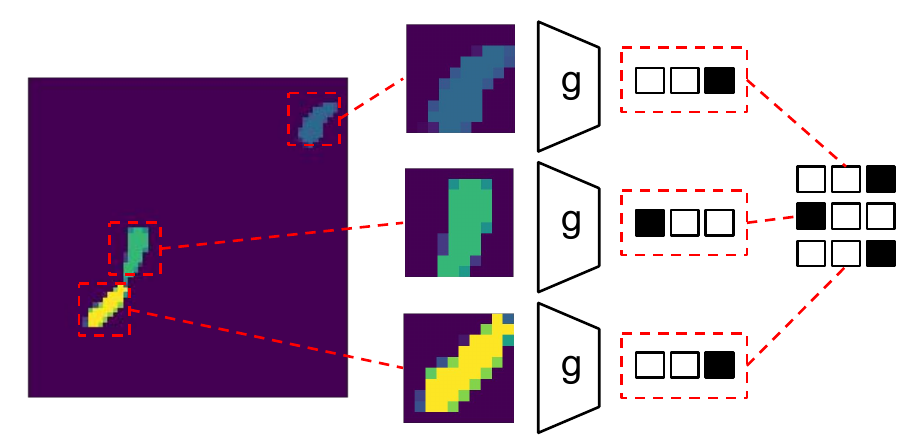}
    \caption{Architecture and sample data used for image and observation conditioned data association prediction.}
    \label{fig:arch}
\end{figure}

As a second illustration of the proposed approach, we explore a vision-based tracking application, where the network for data association is trained to predict permutations conditioned on image patches surrounding measurements. Here we investigate the capacity of a detection model trained by smoothing on one trace to generalise to new settings, using a mice tracking dataset \cite{10.1371/journal.pone.0074557}. Our goal is to learn to identify mice identities given an image (Figure \ref{fig:arch}) and randomly ordered mice position measurements. 

This dataset is significantly larger than those used for previous experiments, so a mini-batch stochastic gradient descent training strategy is used. Here, batches of observation sequences are sampled from the available training data (9988 frames), and gradient descent steps are taken based on the average marginal likelihood across batches.

As shown in Figure \ref{fig:arch}, for each image observation in the sequence, image patches are cropped around measurements, with each of these patches is passed through a fully connected neural network (2 layers of 8 neurons with ReLU activations) with a log softmax output layer ($g_\theta$) of dimension $K$, the number of mice to be tracked. Each of these output vectors is stacked to form a square matrix $X$ and the Sinkhorn operation then applied to produce a permutation matrix. The marginal likelihood is computed by Kalman smoothing \cite{rauch1965maximum} and gradient steps taken to train the network $g_\theta$, following Algorithm \ref{alg} and including graduated optimisation, but averaging across mini-batch sequences. After training, the detector network $g_\theta$ can be used for object recognition in downstream settings.
\begin{table}
    \centering
    \caption{Detection accuracy on held-out test set (10640 frames).    \label{tab:mice_results}}
    \begin{tabular}{l|l|l}
        \hline
        Ground Truth Supervised & Hungarian Supervised & Neural DA+GO\\ 
        \hline
         99.24 \% & 41.30 \% & 99.69 \% \\
         \hline
    \end{tabular}
\end{table}

As baselines, we compare to a network trained using supervised learning and ground truth labelling, alongside an alternative unsupervised learning strategy motivated by \cite{karthik2020simple}, where noisy assignment labels are obtained by applying filtering with the Hungarian algorithm operating on a distance matrix between measurements and tracks. These labels are then paired with images to produce a dataset that is used to train a detector network. Results shown in Table \ref{tab:mice_results} show that jointly training and smoothing with the proposed expectation maximisation approach is highly effective, and performs on par with supervised learning from ground truth labelling in this simplified setting. Supervised learning using Hungarian assignment labels is less effective, as inevitable assignment errors pollute the dataset.

\section{Limitations and future work}

Throughout this work, we have assumed that the number of objects to be tracked remains fixed, and ignored the effects of spurious clutter measurements. The latter could be addressed through the addition of a clutter class.

The joint optimisation task proposed in this work is challenging, which limits the levels of training signal that can propagate through the network to identify more subtle contextual information that assists in data association. Future work exploring the inclusion of pre-trained architectural components and detector elements would be beneficial to allow the application of the approach to more complex settings.

\section{Conclusion}

This letter introduces an expectation maximisation approach to train neural models for data association in multiple-object tracking applications. This self-supervised training mechanism requires no labelling and produces data-association models (or  detector networks for tracking-by-detection) that can be re-used in downstream applications.

\bibliographystyle{IEEEtran}
\bibliography{references}

\end{document}